\documentclass[10pt,twocolumn,letterpage]{article} 
\pdfoutput=1
\usepackage{times}
\usepackage{url} 
\usepackage{latexsym}
\usepackage{hyperref} 
\usepackage{url} 
\usepackage{graphicx}
\usepackage[]{algorithm2e} 
\usepackage{color}
\usepackage{amsmath,amsfonts,amssymb}
\usepackage{authblk}
\newcommand{\RR}{I\!\!R} 
\newcommand{\imagerep}{\lambda} 
\newcommand{\argrep}{\phi} 
\newcommand{\comment}[1]{}
\title{Structured Prediction with Output Embeddings \\for Semantic Image
Annotation}
\author[1]{Ariadna Quattoni\thanks{Corresponding author: ariadna.quattoni@xrce.xerox.com}}
\author[2]{Arnau Ramisa}
\author[3]{Pranava Swaroop Madhyastha}
\author[2]{Edgar Simo-Serra}
\author[2]{Francesc Moreno-Noguer}
\affil[1]{Xerox Research Europe}
\affil[2]{Institut de Rob{\`o}tica i Inform{\`a}tica Industrial, CSIC-UPC, Llorens Artigas 4-6, Barcelona, Spain}
\affil[3]{TALP Research Center, Universitat Polit\`{e}cnica de Catalunya, Campus Nord UPC, Barcelona}
\date{}
\begin{document}
\maketitle
\begin{abstract}
We address the task of annotating images with semantic tuples.
Solving this problem requires an algorithm which is able to deal with hundreds
of classes for each argument of the tuple.  In such contexts, data
sparsity becomes a key challenge, as there will be a large number of
classes for which only a few examples are available. We propose
handling this by incorporating feature representations of both the
inputs (images) and outputs (argument classes) into a factorized
log-linear model, and exploiting the flexibility of scoring functions
based on bilinear forms.  Experiments show that integrating feature
representations of the outputs in the structured prediction model
leads to better overall predictions. We also conclude that the best
output representation is specific for each type of argument.
\end{abstract}
\section{Introduction} \label{ssec:intro}
Many important problems in machine learning can be framed as
structured prediction tasks where the goal is to learn functions that
map inputs to structured outputs such as sequences, trees or general
graphs. A wide range of applications involve learning over large state
spaces, i.e., if the output is a labeled graph, each node of the graph
may take values over a potentially large set of labels. Data sparsity
then becomes a major challenge, as there will be a potentially large number of
classes with  few training examples.

\begin{figure*}[t!]
\begin{center}
\includegraphics[width=0.58\linewidth]{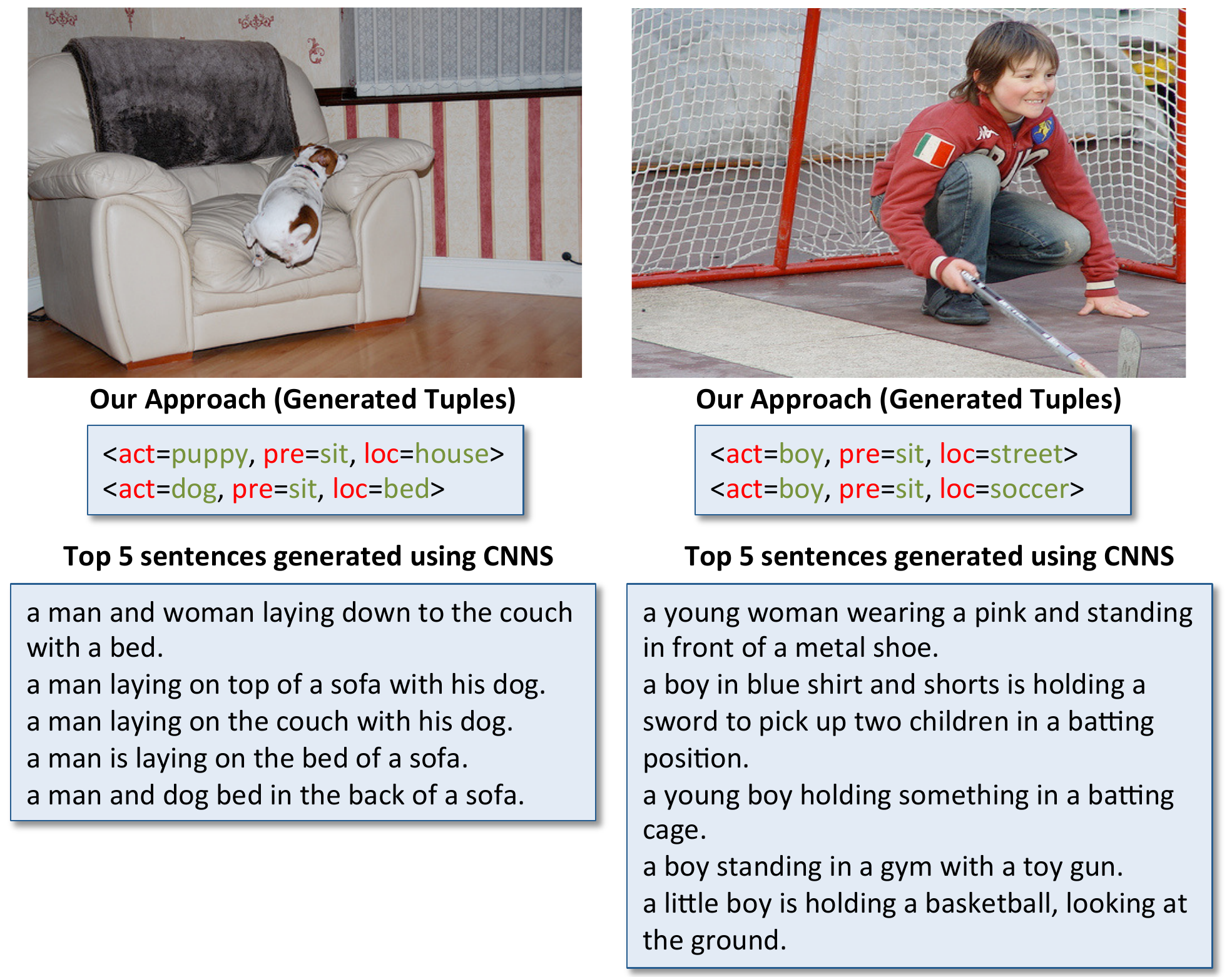}
\end{center}
\vspace{-0.4cm}
\caption{Automatic Tuple Generation. The proposed approach allows generating
semantic tuples that have not been jointly observed before. For instance, in
the left test image,  the joint tuples $\langle puppy,sit,house\rangle$ and
$\langle dog,sit,bed\rangle$ are not present in the training set, but  our
compositional approach can generate them.  
}
\label{fig:intro}
\end{figure*}

Within this context, we are interested in the task of predicting
semantic tuples for images.  That is, given an input image we seek to
predict what are the events or actions ({\em
predicates}), who and what are the participants ({\em
actors}) of the actions and where is the action taking place ({\em locatives}).
Fig.~\ref{fig:intro} shows two examples of the kind
of results we obtain. To handle the data sparsity challenge imposed by the
large state space, we will leverage an approach that has proven to be useful in
multiclass and multilabel prediction tasks~\cite{akata,weston}. The
main idea is to represent a value for an argument $a$ using a feature
vector representation $\argrep \in \RR^{n} $. We will later describe
in more detail the actual representations that we used and how they
are computed but for now imagine that we represent an argument by a
real vector where each component encodes some particular properties of
the argument. We will integrate this argument representation into the
structured prediction framework.

More specifically, we consider standard factorized linear models where
the score of an input/output pair is the sum of the scores, usually
called potentials, of each factor. In our case we will have unary
potentials that measure the compatibility between an image and an
argument of a tuple, and binary potentials that measure the
compatibility between pairs of arguments in a tuple. Typically, both unary and
binary potentials are  linear functions of some feature
representation of the input/output pair.   In contrast, we will consider a
model that exploits bilinear unary potentials $\phi(y,x)$ of the form
$v_{y}^{\intercal}Wx$,  where $v_y \in \RR^{n}$ is some real vector
representation of an argument $l \in L$ and $x \in \RR^d$ is a $d$
dimensional feature representation of an image.  Similarly, the binary
potentials $\alpha(y,y')$ will be of the form $v_{y}^{\intercal} Z {v_y'}$
for a pair of arguments $(y, y')$.  The rank of $W$ and $Z$ can be
interpreted as the intrinsic dimensionality of a low-dimensional
embedding of the inputs and arguments feature representation.  Thus, if
we want computationally efficient models (i.e. few features) it is
natural to use the rank of $W$ and $Z$ as a complexity penalty. Since
using the rank would lead to a non-convex problem, we  use instead
the nuclear norm as a convex relaxation. We conduct experiments with two
different feature representations of
the outputs and show that integrating an output feature representation
in the structured prediction model leads to better overall
predictions. We also conclude from our results that the best output
representation is different for each argument type.

\begin{figure}[t!]
\begin{center}
\includegraphics[width=1.0\linewidth, height=5.3cm]{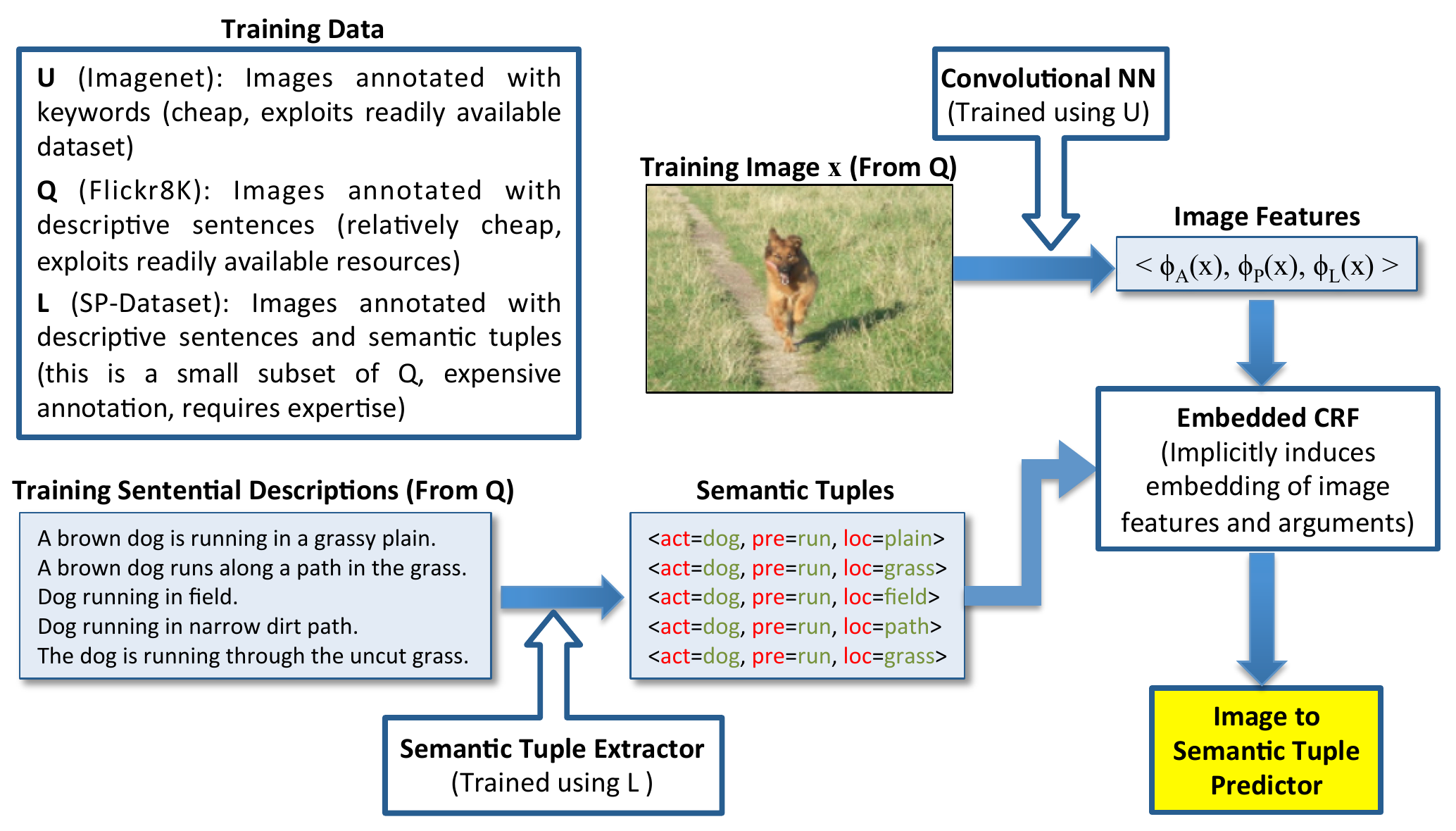}
\end{center}
\vspace{-5mm}
\caption{Overview of our approach.}
\label{fig:overview}
\vspace{-2mm}
\end{figure}

\section{Semantic Tuple Image Annotation}
\subsection{Task} We will address the task of predicting semantic tuples for
images. Following~\cite{FarhadiECCV2010}, we will focus on a simple semantic
representation that considers three basic arguments: predicate, actors
and locative. 
For example, an image might be annotated with the semantic tuples: $\langle
run, dog, park \rangle$ and
$\langle play, dog, grass \rangle$. We call each field of a tuple an {\em
argument}. For example, in the tuple $t =\langle play, dog, grass \rangle$,
``$play$'' is the  argument of the predicate field, ``$dog$'' is the actor and
``$grass$''  the argument of the locative field.

Given this representation, we can formally define our problem as that of
learning a function $\theta:~X~\times~P~\times~A~\times~L \to \RR$ that scores
the compatibility between images and semantic tuples. Here, $X$ is the space of
images, $P$ is a discrete set of predicate arguments, $A$ is a set of actor
arguments and $L$ is a set of locative arguments. We are particularly
interested in cases where $|P|$, $|A|$ and $|L|$ are reasonably large. We will
use $T=P \times A \times L$ to refer to the set of possible tuples, and denote
by $\langle p \ a \ l \rangle$ a specific instance of the tuple. To learn this
function we are provided with a training set $Q$.  Each example
in this set consists of an image $x$ and a set of corresponding semantic tuples
$\{t_c\}$ which describe the events occurring in the image. Our goal is to use
$Q$ to learn a model for the conditional probability of a tuple given and
image. We will use this model to predict semantic tuples for test images by
computing the tuples that have highest conditional probability according to our
learnt model.

\subsection{Dataset}
\begin{figure*}[t!]
\begin{center}
\includegraphics*[width=\textwidth]{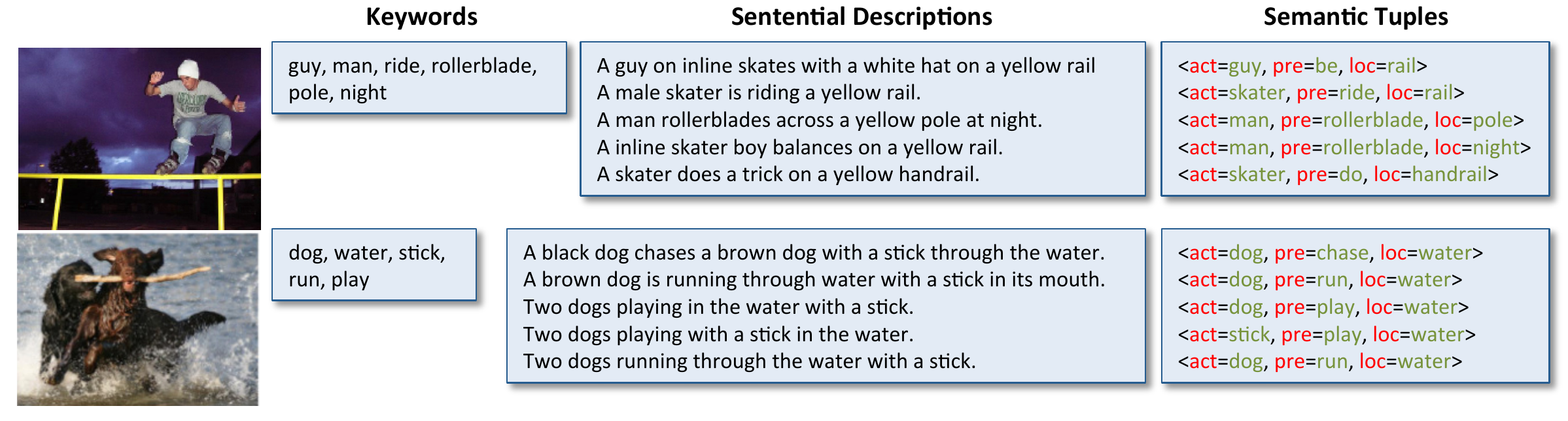}
\end{center}
\caption{Sample images, keywords, sentences and semantic tuples from the
augmented Flickr-8K dataset.}
\label{fig:examples_dataset}
\end{figure*}

While some datasets of images associated with semantic tuples are already
available \cite{FarhadiECCV2010},
they only consider small state spaces for each argument type. To address this
limitation we decided to create
a new dataset of images annotated with semantic tuples. In contrast to previous
datasets, we consider a more
realistic range of possible argument values. In addition, our dataset has the
advantage that every image is annotated
with both the underlying semantics in the form of semantic tuples and natural
language captions that constitute different
lexical realizations of the same underlying semantics. To create our dataset we
used a subset of the Flickr8k dataset, proposed in
Hodosh et al.~\cite{HodoshYH13}. This dataset consists of 8,000 images taken
from
Flickr of people and animals performing some action, with five
crowd-sourced descriptive captions for each one. These captions are sought to
be concrete descriptions of what can be seen in the image rather than abstract
or conceptual descriptions of non-visible elements (e.g., people or street
names, or the mood of the image). This type of language is also known as Visually 
Descriptive Language~\cite{gaizauskas2015}.

We asked human annotators to annotate 1,544 image captions, corresponding to 311 images
(approximately one third of the development set), producing more than 2,000
semantic tuples of predicates, actors and locatives. Annotators were required to
annotate every caption with their corresponding
semantic tuples without looking at the referent image. We do this to ensure an
alignment between the
information contained in the captions and their corresponding semantic tuples. 
Captions are annotated with tuples that consist of a predicate, a patient, an
agent and a locative (indeed the patient, the agent and the locative could
themselves consist of multiple arguments but for simplicity we regard them as
single arguments). For example, the caption ``\emph{A brown dog is playing and
holding a ball in a crowded park}'' will have the associated
tuples: $\langle \ predicate=play, \ agent=dog, \ pacient=null, \ locative=park
\rangle$ and  $\langle \ predicate=hold, \ agent=dog, \ pacient=ball, \
locative= park \rangle$. Notice that while these annotations are similar to
PropBank style semantic role annotations, there are also some differences.
First, we do not annotate atomic sentences but captions that might actually
consist of multiple sentences. Second, the annotation is done at the highest
semantic level and annotators are allowed to make logical inferences to resolve
the arguments of a predicate. For example we would annotate the caption:
``\emph{A man is standing on the street. He is holding a camera}'' with
$\langle \ predicate=standing, \ agent=man, \ pacient=null, \ locative= street
\rangle$ and  $\langle \ predicate=hold, \ agent=man, \ pacient=null, \
locative= street \rangle$.
Figure~\ref{fig:examples_dataset} shows two sample images with captions and
annotated semantic tuples.
For the experiments we partitioned the set of 311 images (and their
corresponding captions and tuples) into a training set of 150 images,
a validation set of 50 (used to adjust parameters) and a test set of
100 images.

To enlarge the manually annotated dataset we first used the data of captions
paired with semantic tuples to train a model that can predict semantic tuples
from image captions. Similar to previous work we start by computing several
linguistic features of the captions,  ranging from shallow part of speech tags
to dependency parsing and semantic role labeling~\footnote{We use the
linguistic analyzer of~\cite{freeling}}. We extract the predicates by looking
at the words tagged as verbs by the POS tagger. Then, the extraction of
arguments for each predicate is resolved as a classification problem. More
specifically, for each detected predicate in a sentence we regard each
noun as a positive or negative training example of a given relation depending
on whether the candidate noun is or is not an argument of the predicate. We use
these examples to train a discriminative classifier that decides if a candidate
noun is or is not an argument of a given predicate in a given sentence. This
classifier exploits several  linguistic features computed over the syntactic
path of the dependency tree connecting the candidate noun and the predicate. 
As a classifier we trained a linear SVM. We run the learnt tuple predictor
model on all the remaining 6,000 training images and corresponding captions of
the Ficker8k
dataset and produced a larger dataset of images paired with semantic tuples
\footnote{In the experimental section we actually build models to predict
coarser triplets that consist of a locative a predicate and an actor.
To convert from the finer $\langle predicate, \ agent , \  patient,  \ locative
\rangle$ annotations to the coarser annotations $ \langle
predicate, \ actor, \ locative \rangle$ we simply map the finer annotation to
two coarser
tuple annotations, one tuple for the actor and one tuple for the patient. }.
\section{Incorporating Output Feature Representations into a Factorized Linear
Model} \label{ssec:techical}

For simplicity we will consider factorized sequence models over sequences of
fixed length. However,  all the ideas
we present  can be easily generalized to other structured prediction
settings. In this section we first describe the general model and learning
algorithm (Sections~\ref{sec:bilinearmodels} and~\ref{sec:learning},
respectively), and then, in Section~\ref{sec:bilinearfortuples}, we focus on
the specific problem of learning tuples given input images.

\subsection{Bilinear Models with Output Feature Representations}
\label{sec:bilinearmodels}
Let $x$ be an input, and let $y= [y_1 \ldots y_T]$ be some output sequence
 where $y_i \in L$ for some
set of states $L$. We are interested in learning a
model that computes $P(y|x)$, i.e. the conditional probability of a sequence
$y$ given some input $x$.  We will consider CRF-like factorized log-linear
models that
take the form:
\begin{equation}
   P(y|x)= \frac{\exp{\theta(x,y)}} {\sum_{y} \exp{\theta(x,y)}}
\end{equation}

The scoring function $\theta(x,y)$ is modeled as a sum of unary and binary
bilinear potentials and is defined as:
\begin{equation}
   \theta(x,y) = \sum_{t=1}^{T} {v_{y_t}^{\intercal} W_{t} \phi(x,t)}
               + \sum_{t=1}^{T} {v_{y_t}^{\intercal} Z_{t} v_{y_{t+1}}}
\end{equation}
\noindent where $v_{y} \in \RR ^{|n|}$ is a feature representation of label $y \in L$,
and $\phi(x,t) \in \RR^{d}$ is a feature representation of the $t$-th input
factor of $x$. 

The first set of terms in the above equation are usually refered as unary
potentials and measure the compatibility between a single state at $t$ and the
feature representation of input factor $t$. The second set of terms are the
binary potentials and measure the compatibility between pairs of states at
adjacent factors. The scoring  function $\theta(x,y)$ is fully parameterized by
the unary parameter matrices $W \in \RR^{|n| \times d}$ and the binary
parameter matrices $ Z \in \RR^{|n| \times |n|} $. 

We will later describe the actual label feature representations that we used in
our experiments. But for now, it suffices to say that the main idea is to
define a feature space so that semantically similar labels will be close in
that space. Like in the multilabel scenario~\cite{akata,weston}, having full
feature representations for arguments will allow us to share information across
different classes.

One of the most important advantages of using feature representations for the
outputs is that they give us the ability to  generalize better. This is because
with a good output feature representation, our model should be able to make
sensible predictions about pairs of arguments that were not observed at
training. This is easy to see: consider a case were we have a pair of arguments
represented with feature vectors $a_1$ and  $a_2$ and suppose that we have not
observed the factor $a_1,a_2$ in our training data but we have observed the
factor $b_1,b_2$.  Then if $a_1$ is close in the feature space to argument
$b_1$ and $a_2$ is close to $b_2$ our model will predict that $a_1$ and $a_2$
are compatible. That is, it will assign probability to the pair of arguments
$a_1,a_2$
which seems a natural generalization from the observed training data. 

This kind of representation also has interesting interpretations in terms of
the ranks of $W$ and $Z$. Let 
$W = U \Sigma V$ be the singular value decomposition of $W$. We can then write the
unary potential  $v_{y}^{\intercal} W \phi(x,t)$ as: 
\begin{equation}
 v_{y}^{\intercal}U \ \Sigma \ [V \phi(x,t)].
\end{equation}
Thus, we can regard the bilinear form as a function computing a weighted inner
product between some real embedding $v_{y}^{\intercal}U$ representing state $y$, and some
real embedding $[V \phi(x,t)]$ representing input factor $t$. The rank of $W$
gives us the intrinsic dimensionality of the embedding. Therefore, if we seek to
induce shared low-dimensional embeddings across different states it seems
reasonable to impose a low rank penalty on $W$. 

Similarly, let $Z=U \Sigma V$
be  the singular value decomposition of $Z$. We can write the binary
potentials $v_{y}^{\intercal} Z v_{y'}$ as: 
\begin{equation}
v_{y}^{\intercal}U \ \Sigma \ V v_{y'}
\end{equation}
and thus the binary potentials compute a weighted inner product between a real
embedding of state $y$ and a real embedding of state $y'$. Again, the rank
of $Z$ gives us the intrinsic dimensionality of the embedding and, to induce a
low dimensional embedding for binary potentials, we will impose a low
rank penalty on $Z$. In practice, imposing low-rank constraints, would lead to a
hard optimization problem, so instead we will use the nuclear norm as a convex
relaxation of the rank function.
\subsection{Learning Algorithm}
\label{sec:learning}
\begin{algorithm}[t!] 
    \SetAlgoLined
    Inputs: $D, \eta,\gamma, c$ \\ 
    Output: $W$ \\
    Initialize $W = 0$ \\ 
    \While{$t \leq MaxIter$}{ 
        $G_t = \partial(Loss(D,\{W\}))/\partial{W}$;\\  
    $W_{t+0.5} = W_{t}-\nu_{t} G_{t}$; \tcp{$\nu_t$ is the learning rate}
    $W_{t+0.5} = U \Sigma V^{\intercal}$;\\
    $\forall$ unary potentials define a diagonal matrix $\Sigma'$ 
    such that: $\sigma_i'=\max[\sigma_i-\nu_t\eta]$;\\
    $W_{t+1}=U\Sigma'V^{\intercal}$;\\
    $\forall$ binary potentials define a diagonal matrix $\Sigma'$
    such that: $\sigma_i'=\max[\sigma_i-\nu_t\gamma]$;\\
    $W_{t+1}=U\Sigma'V^{\intercal}$;
    }%
    \caption{Learning Algorithm}
    \label{fig:algorithm} 
\end{algorithm}

After having  described the type of scoring functions we are interested in, we
now turn our attention to the learning problem. That is, given a training set
$D= \{ \langle x \ y \rangle \}$ of pairs of inputs $x$ and output sequences 
$y$ we need to learn the parameters $ \{W\} $ and $\{Z \}$. For this purpose we
will do standard max-likelihood estimation and find the parameters that
minimize the conditional negative log-likelihood of the data in $D$. That is, we
will find the $\{W\}$ and $\{Z\}$ that minimize the following loss function
$Loss(D,\{W\}, \{Z\} $):
\begin{equation}
   -\sum_{\langle x \ y \rangle \in D } \log P(y|x; \{W\}, \{Z\})
\label{eq:loss}
\nonumber
\end{equation}
It can be shown that this loss function is convex on $\{ W \}$ and $\{ Z
\}$ whenever $\theta(x,t; \{ W \}, \{ Z \} )$ is convex, which is the case for
our scoring function. 

Recall that we are interested in learning low-rank unary and binary potentials.
To this end we follow the standard approach which is to use   the nuclear
norm $|W|_{*}$ and $|Z|_{*}$ (i.e. the $l_1$ norm of the singular values)  as a
convex approximation of the rank function. Putting all this together, the final
optimization problem becomes:
\begin{equation}
   \min_{\{W\}} Loss(D,\{W\}) + c_1
      \sum_{t} |W_{t}|_{*} + c_2 \sum_{t} |Z_{t}|_{*}
   \label{eq:optimization}
\end{equation}
\noindent where $Loss(D,\{W\})=\sum_{d \in D} Loss(d,\{W\})$ is
the negative log likelihood function and $c_1$ and $c_2$ are two
constants that control the trade off between minimizing the loss and the
implicit dimensionality of the embeddings. 

In recent years, many algorithms have been proposed for optimizing trace norm
regularized problems (e.g., see \cite{JaggiS10,Shalev-ShwartzSSC11,JiICML2009}).
We use a simple optimization scheme known as
Forward Backward Splitting, or FOBOS~\cite{DuchiJMLR2009}. It can be
shown that FOBOS converges to the global optimum at a
$O(1/\epsilon^{2})$ rate.

The main steps of the optimization involve computing the gradient of the loss
function and performing singular value decomposition on each $W$ and $Z$. In
our case, computing the gradient involves computing marginal probabilities for
unary and binary potentials which has a cost of $O(|L|^2)$ and the cost of the
SVD computation for each $W$ in $\{ W \}$ and each $Z$  in $\{ Z\}$.

\subsection{Bilinear CRF for Predicate Prediction}
\label{sec:bilinearfortuples}
For our task we will consider a simple factorized scoring function 
$\theta(x,\langle p \ a \ l \rangle)$ that has unary terms relating arguments
of the same kind, and binary factors associated with the
$locative-predicate$ pair and  with the $predicate-actor$
pair. Since this corresponds to a chain structure, $\arg\max_{t \in T}
\theta(x;\langle p \ a \ l \rangle )$ can be efficiently computed using Viterbi
decoding in time $O(N^2)$, where $N = \max(|P|,|A|,|L|)$. Similarly, we can also
find the top $k$ predictions in $O(kN^2)$. Alternatively, we could have defined
the relationship between arguments via  a fully connected graph and use
approximate inference methods. 

More specifically,  the scoring function of the bilinear CRF we contemplate takes  the form:
\begin{eqnarray}
   \theta(x,\langle p \ a \ l \rangle)  &=&
   \imagerep_{loc}(l)^{\intercal} W_{loc} \argrep_{loc}(l) \nonumber \\ && +
   \imagerep_{pre}(p)^{\intercal} W_{pre} \argrep_{pre}(p)  \nonumber \\ && +
   \imagerep_{act}(a)^{\intercal} W_{act} \argrep_{act}(a)     \nonumber \\ && +
   \argrep_{loc}(l)^{\intercal} W^{loc}_{pre} \argrep_{pre}(p) \nonumber \\ && +
   \argrep_{pre}(p)^{\intercal} W^{pre}_{act} \argrep_{act}(a)
\label{eq:scoringfunction}
\end{eqnarray}
where the $\imagerep$'s are the image representations and the $\argrep$'s the textual ones. 
The unary potentials (first three terms in Eq.~\ref{eq:scoringfunction})
measure the compatibility between image and  
semantic arguments; the first binary potential measures the
compatibility between the semantic representations of locatives and
predicates, and the second
binary potential measures the compatibility between   predicates and 
actors. The scoring function is fully parameterized by the unary
parameter matrices $W_{loc} \in \RR^{d \times nl}$, $W_{pre} \in
\RR^{d \times np}$ and $W_{a} \in \RR^{d \times na}$ and by the binary
parameter matrices $W^{loc}_{pre} \in \RR^{nl \times np}$ and
$W^{pre}_{act} \in \RR^{np \times na}$.  The parameters  $nl$, $np$ and $na$ are the
dimensionalities of  feature representations for the locatives, predicates and actors.

Note that if we let the argument representation $\argrep(r)$ be an
indicator vector in $\RR^{|L|}$ 
we obtain the usual parametrization of a standard
factorized linear model:
\begin{eqnarray}
   \theta(x,\langle p \ a \ l \rangle)
    &=& \imagerep_{loc}(l)^{\intercal} w_{loc}^{l}  \nonumber \\ && +
    \imagerep_{pre}(p)^{\intercal} w_{pre}^{p}   \nonumber \\ && +
    \imagerep_{act}(a)^{\intercal} w_{act}^{a}   \nonumber \\ & & + W^{loc}_{pred}(l,p)
+ W^{pred}_{act}(p,a)  \nonumber 
\end{eqnarray}

Like in the multilabel scenario~\cite{akata,weston}, having full feature
representations for arguments instead of indicator vectors will allow us to
share information across different classes. In fact, we will use the model that
uses indicator vectors as a baseline in our experiments.

\section{Representing Semantic Arguments}
\label{argrep}

Recall that in order to handle the large number of possible arguments per field 
(i.e.~data sparsity) our model assumes the existence of some feature representation
for each argument and type $\argrep_{pred}(p) \in \RR^{np}$, $\argrep_{act}(a)
\in \RR^{na}$ and $\argrep_{loc} \in \RR^{nl}$.  It is then that by learning an
embedding of these vectors we will be able to share information across
different classes. Intuitively, the feature vectors should describe properties
of the arguments and should be defined so that  feature vectors that are close
to each other represent arguments that are semantically similar.

We will conduct experiments with two different feature
representations: 1) Fully unsupervised \textit{Skip-Gram based Continuous Word
Representations} (SCWR)  and 2) a feature representation
computed using the $\langle caption, semantic-tuples \rangle$ pairs,
that we call \textit{Semantic Equivalence Representation} (SER). We next
describe in more detail each of these representations.

\subsection{Semantic Equivalence Representation}

We want to exploit the dataset of captions paired with semantic
tuples to induce a useful feature representation for arguments. For this we
will propose a way to illustrate the fact that any pair of
semantic tuples associated with the same image will likely be 
describing the same event.  Thus, they are in essence different ways
of lexicalizing the same underlying concept.

Let's look at a concrete example. Imagine that we have an image annotated with
the tuples: $\langle play,\ dog, \ water \rangle$ and 
$\langle play, \ dog, \ river\rangle$. Since both tuples describe the same image, 
it is quite likely that both ``$river$'' and ``$water$'' refer to the
same real world entity, i.e, ``$river$'' and ``$water$'' are
'semantically equivalent' for this image. Using this idea we build
a representation $\argrep_{loc}(i) \in \RR^{|L|}$ where the $j$-th
dimension corresponds to the number of times the argument $j$ has been
semantically equivalent to argument $i$.

More precisely, we compute the probability that argument $j$ can be exchanged
with argument $i$ as: $\frac{[ i, j]_{sr}}{ \sum_j [i, j]_{sr}}$, where
$[i,j]_{sr}$ is the number of times that $i$ and $j$ have appeared as
annotations of the same image and with the same other arguments. For example,
for the actor arguments $[i,j]_{sr}$ represents the number of time that actor $i$
and actor $j$ have appeared with the same locative and predicate as
descriptions of the same image. Here is  a concrete example of the feature
vector for the locative `water' (we report the
non-zero dimensions and their corresponding value): $\argrep_{loc}(water)$\textit{=[
air 0.03, beach 0.06, boat 0.03, canoe 0.03, dock 0.13, grass 0.06, kayak
0.06, lake 0.06, mud 0.03, ocean 0.16, platform 0.03, pond 0.06, puddle
0.1, rock 0.03, snow 0.03, tree 0.03, waterfall 0.03]}. Thus, according to
the computed representation, `water' is semantically most similar to `ocean'.

\subsection{Skip-Gram based Continuous Word Representations}
Recently, there has been interest  in learning
word-representations, which have been proven to be useful for many structure
prediction tasks~\cite{Turian2010,Koo08,Tackstrom2012}. We use
continuous word representations (also known as distributed representations) to
tailor a task-specific embedding. Continuous word representations consist of 
neural network-based low-dimensional real valued vectors of each word. We use
\cite{Mikolov2013}'s skip-gram based approach for inducing continuous word
representations. Skip-gram based representations are essentially a single layer
neural network, and are based on inner products between two word vectors.
The objective function in a skip-gram is to predict a word's context given the
word itself. We
use the trained continuous word representations computed over the Google News
dataset(100 billion words), that is publicly
available\footnote{https://code.google.com/p/word2vec/}, in our experiments.

\section{Related Work}
\label{ssec:related}

In recent years, some works have tackled the problem of generating rich textual
descriptions of images. One of the pioneers is \cite{KulkarniCVPR2011}, where a
CRF model combines the output of several vision systems to produce input for a
language generation method. This seminal work, however, only considered a
limited set of a few tens of labels, while we aim at dealing with potentially
hundreds of labels simultaneously. In \cite{FarhadiECCV2010}, the authors find
the similarity between sentences and images in a ``meaning'' space, represented
by semantic tuples which are very similar to ours: triplets of \emph{object},
\emph{action} and \emph{scene}. The main difference  with this work is that
it uses a ruled based system to extract semantic tuples from dependency trees
where we train a model that predicts semantic tuples and, most importantly, it
uses a standard factorized linear model while we propose a model that leverages
feature representations of arguments, and can therefore handle significantly larger
state spaces.

Other works focus on the simplified problem  of ranking  human-generated
captions for images. In \cite{HodoshYH13} the authors propose to use
Kernel Canonical Correlation Analysis to project images and their
captions into a joint representation space, in which images and
captions can be related and ranked to perform illustration and
annotation tasks. However, the system cannot be used to generate novel
image descriptions for new images and, since a kernel is necessary, it
has limitations on the number of image/caption pairs that can be used
to define the subspace. In a follow-up work, the authors address
improving the text/image embeddings with abundant weakly-annotated
data from Flickr and similar sites using a stacked representation
\cite{gong14eccv}. To cope with the large amounts of data, Normalized
Canonical Correlation Analysis is used. Socher et
al.~\cite{SocherTACL2014} also address the ranking of images given a
sentence and vice-versa using a common subspace, also known as zero-shot
learning. Recursive Neural
Networks are used to learn this common representation. The work of
\cite{KuznetsovaTACL2014}  performs natural text
generation from images using a bank of detectors to find objects and
compressing the text to retrieve `generalizable' small fragments. On top of
this,  a tree approach is used to construct sentences given the observations
and fragments. However, the  sentences produced this way can be easily
corrupted by wrongly retrieved segments. 

Recent works use deep networks to address the problem:
\cite{VinyalsCVPR2015} propose a pure deep network approach, where
convolutional neural networks are used both to extract image features
and recursive deep network to generate the text. The system is trained
to maximize likelihood end-to-end. \cite{KarpathyCVPR2015} use a
common multi-modal embedding to align text and images, and a recurrent
neural network is trained to generate sentences directly from the
image pixels. Although these methods report good results in terms of BLEU score 
agreement with gold captions, they do not model the underlying visual
predicates which is the goal of this paper.

Using label embeddings and its combination with bilinear forms has been
previously proposed in the context of multiclass and multilabel image
classification~\cite{akata,weston}, but to the best of our knowledge there is
no previous work on leveraging output embeddings in the context of structured
prediction. Thus, besides the concrete application to semantic tuple image
generation, this paper presents a useful modeling tool for handling structured
prediction problems in large state spaces. Our model can be used whenever we
have some means of computing a feature representation of the outputs.
\section{Experiments}
\begin{figure*}[t!]
\begin{center}
\includegraphics[width=1\linewidth, height=7.8cm]{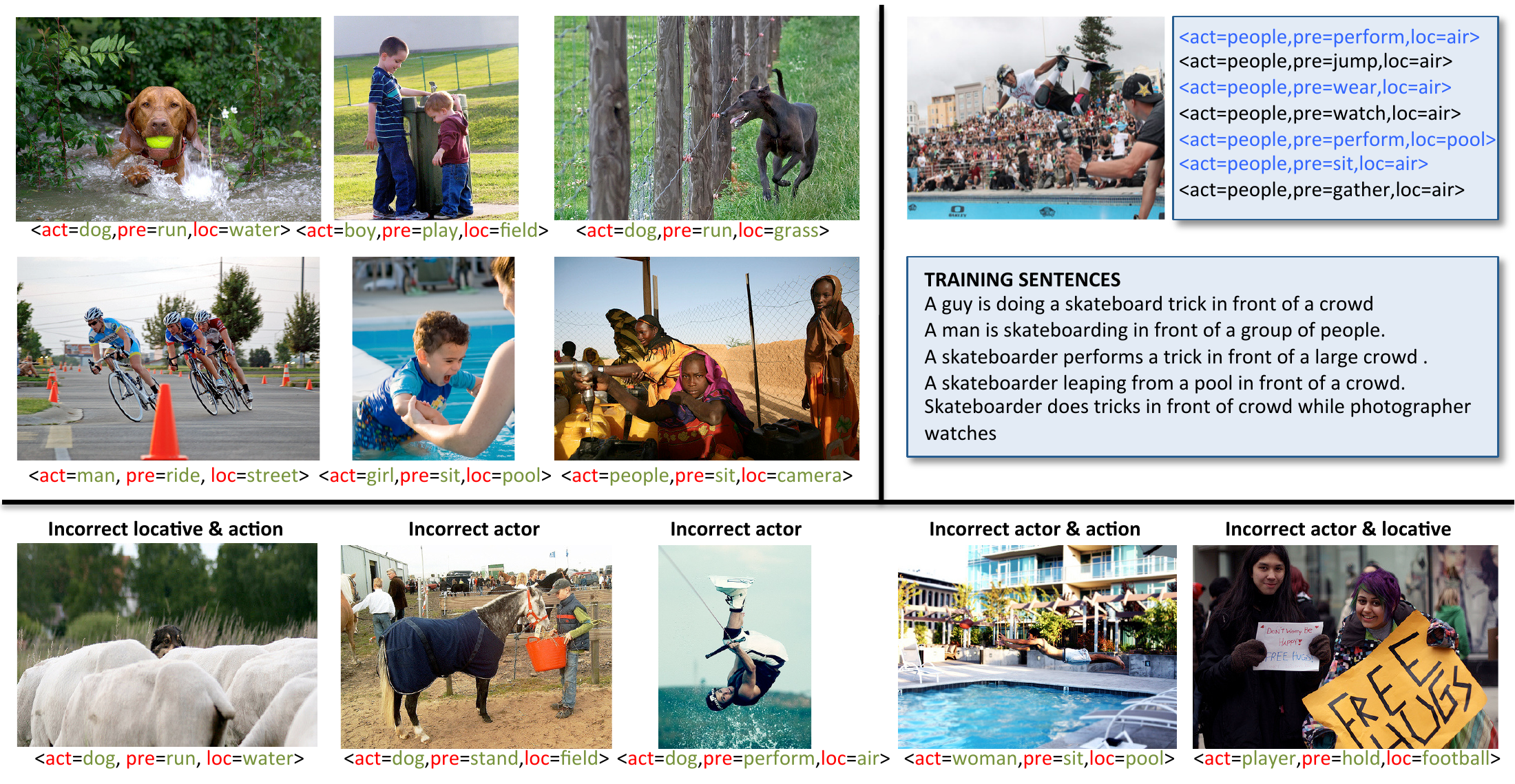}
\end{center}
\vspace{-0.7cm}
\caption{Samples of  predicted tuples. {\bf Top-left:} Examples of  visually correct predictions. {\bf Bottom:} Typical errors on one or several arguments. {\bf Top-right:} Sample image and its top predicted tuples.  The tuples in blue were not observed neither in the SP-Dataset nor in the automatically enlarged dataset. Note that all of them are descriptive of what is occurring in the scene.  }
\label{fig:qualitativeresults}
\vspace{-2mm}
\end{figure*}

\begin{figure*}[ht!]
   \includegraphics[width=0.5\textwidth]{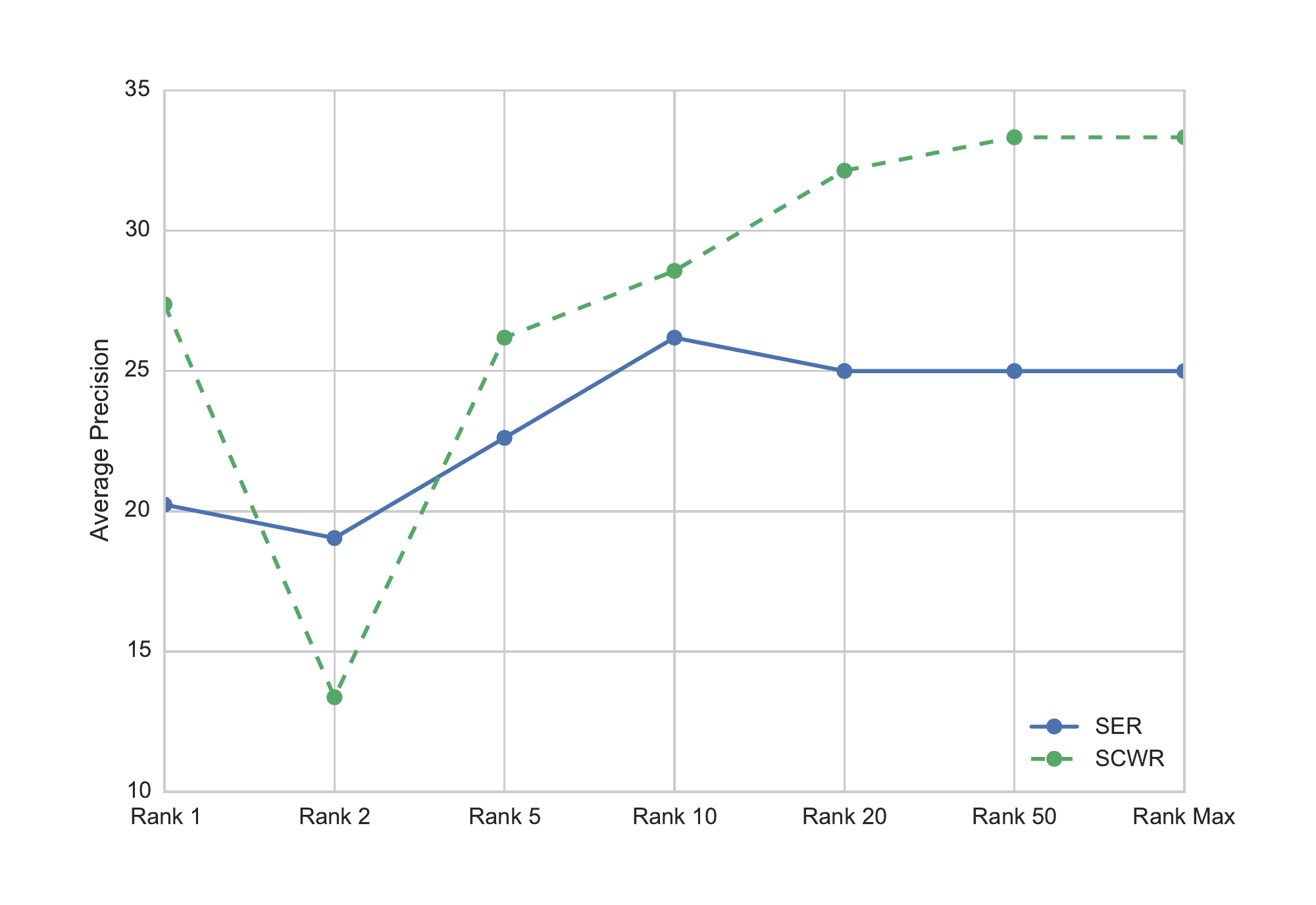}
   \includegraphics[width=0.5\textwidth]{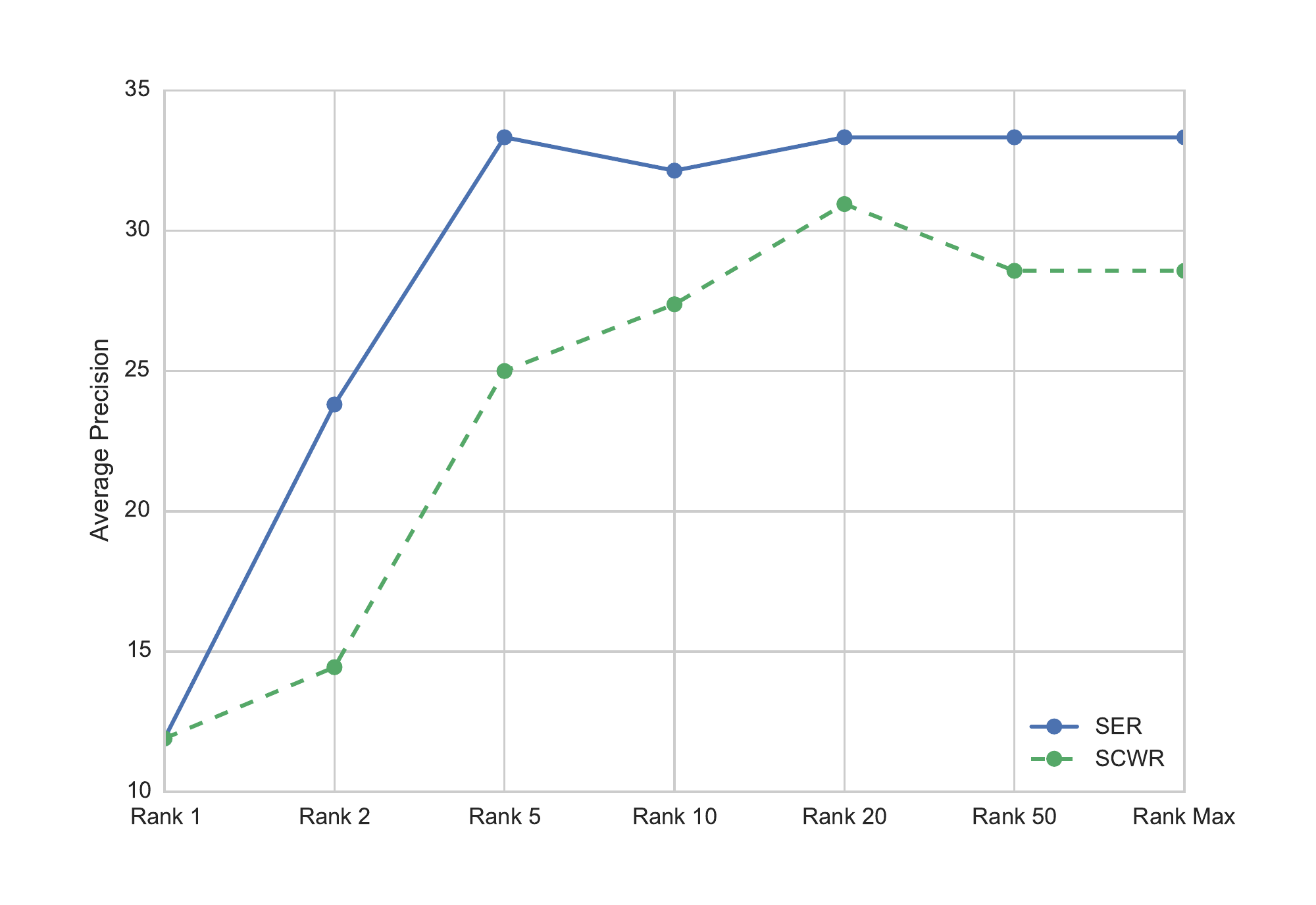}
   \caption{Performance as a function of the size of the intrinsic embedded space
   for predicate (left) and locative (right) arguments.}
   \label{fig:curves}
\end{figure*}

As it is standard practice, in order to compute  image representations ($\imagerep$-vectors in Eq.\ref{eq:scoringfunction}), we   use the 4,096-dimensional second to
last layer of a Convolutional Neural Network (CNN). The full network has 5 convolutional layers followed by 3 fully
connected layers,  and obtained the best performance in the ILSVRC-2012
challenge.  The network is trained on a subset of ImageNet~\cite{DengCVPR2009}
to classify 1,000 different classes and we use the publicly available
implementation and pre-trained model provided by~\cite{Jia13Caffe}. The
features obtained with this procedure have been shown to generalize well and
outperform traditional hand-crafted features, thus they are already being used
in a wide diversity of tasks~\cite{SocherTACL2014,XuCVPR2014}.

To test our method we used the 100 test images that were annotated with
ground-truth semantic tuples. For locatives, predicates and actors we consider
the 400 most frequent. To measure performance we first compute the top 5 tuples for each image. Then, we define the set of predicted
locatives to be the union of all predicted locatives and we do the same for the other
argument types. Finally, we compute the precision for each type, for example, for the locatives
this is the percentage of predicted locatives that were present in the gold tuples for the corresponding
image. 

The regularization parameters of each model were set using the validation set. We compare the performance of several models:

\begin{itemize}
\item Baseline KCCA:  This model implements the Kernel Canonical Correlation Analysis approach of \cite{HodoshYH13}.
We first note that this approach is able to rank a list of candidate captions but cannot directly generate tuples. To generate tuples 
for test images we first find the caption in the training set that has the highest ranking score for that image and then extract the corresponding
semantic tuples from that caption. These are the tuples that we consider as predictions of the KCCA model.
\item Baseline Separate Predictors (SPred):  We also consider a baseline made of independent predictors for each argument type. 
More specifically we train one-vs-all SVMs (we also tried multi-class SVMs but they did not improve performance) to independently predict locatives,
predicates and actors. For each argument type and candidate label we have a score computed by the corresponding SVM. Given an image we generate the top
tuples that maximize the sum of scores for each argument type.
\item Embedded CRF with Indicator Features (IND), this is a standard factorized log-linear
      model that does not use any feature representation for the outputs.
 \item Embedded CRF with a model that uses the skip-gram continuous word representation of
    outputs (SCWR).
 \item Embedded CRF with a model that uses that semantic equivalence representation of outputs
    (SER).
 \item A combined model that makes predictions using the best feature
    representation for each argument type (COMBO).
\end{itemize}

Table~\ref{tbl:results} reports the results for the baselines and of
the different CRF schemes. The first observation is that the best
performing output feature representation is different for each
argument type. For the locatives the best representation is SER, for
the predicates is the SCWR and for the actors using an output feature
representation causes a drop in performance.  The largest improvement
from using an output feature representation that we obtain is on the
predicate arguments, where we improve almost by 10\% over the
indicator representation by using the skip-gram
representation. Overall, the model that uses the best representation
performs better than the indicator baseline.

Finally, Figure~\ref{fig:curves} shows performance as a function of the dimensionality
of the learnt embedding, i.e. rank of parameter matrices, as we can see the
learnt models are efficient in the sense that they can work well with
low-dimensional projections of the features. 

\begin{small}
\begin{table}[tb]
\begin{center}
\resizebox{\columnwidth}{!}{%
   \begin{tabular}{lllllll}
        & \small{Spred} & \small{KCCA} & \small{IND} & \small{SCWR} & \small{SER} & \small{COMBO} \\
       \hline \small{LOC} & 15 & 23 & 32 & 28 & \bf 33  \\
       \small{PRED} & 11 & 20 & 24 & \bf 33 & 25  \\
       \small{ACT} & 30 & 25 & \bf 52 & 51 & 50  \\
      \hline \small{MEAN} & 18.6 & 22.6 & 36 & 37.3 & 36 & \bf 39.3 \\
      \hline
   \end{tabular}}
\end{center}
\caption{Precision of baseline and CRFs with different output embeddings.}
\label{tbl:results}
\end{table}
\end{small}

\section{Conclusion}
In this paper we have presented a model for exploiting input and
output embeddings in the context of structured prediction.  We have
applied this framework to the problem of predicting compositional
semantic descriptions of images. Our results show the advantages of
using output embeddings for handling large state spaces. We have also
seen that regularizing with the nuclear norm we can obtain
computationally efficient low-rank models with comparable performance.

\section*{Acknowledgments}
This work was partly funded by the Spanish MINECO projects RobInstruct
TIN2014-58178-R, SKATER TIN2012-38584-C06-01, and by the ERA-net
CHISTERA project VISEN PCIN-2013-047.

\begin{thebibliography}{10}\itemsep=-1pt

\bibitem{akata}
Z.~Akata, F.~Perronnin, Z.~Harchaoui, and C.~Schmid.
\newblock Label-embedding for attribute-based classification.
\newblock In {\em Proc. IEEE Conference on Computer Vision and Pattern
  Recognition (CVPR)}, 2013.

\bibitem{DengCVPR2009}
J.~Deng, W.~Dong, R.~Socher, L.-J. Li, K.~Li, and L.~Fei-Fei.
\newblock {ImageNet: A Large-Scale Hierarchical Image Database}.
\newblock In {\em Proc. IEEE Conference on Computer Vision and Pattern
  Recognition (CVPR)}, 2009.

\bibitem{DuchiJMLR2009}
J.~Duchi and Y.~Singer.
\newblock Efficient online and batch learning using forward backward splitting.
\newblock {\em Journal of Machine Learning Research (JMLR)}, 10:2899--2934,
  2009.

\bibitem{FarhadiECCV2010}
A.~Farhadi, M.~Hejrati, M.~Sadeghi, P.~Young, C.~Rashtchian, J.~Hockenmaier,
  and D.~Forsyth.
\newblock Every picture tells a story: Generating sentences from images.
\newblock In {\em Proc. European Conference on Computer Vision (ECCV)}. 2010.

\bibitem{gaizauskas2015}
R.~Gaizauskas, J.~Wang, and A.~Ramisa.
\newblock Defining visually descriptive language.
\newblock In {\em Proceedings of the 2015 Workshop on Vision and Language
  (VL'15): Vision and Language Integration Meets Cognitive Systems}, 2015.

\bibitem{gong14eccv}
Y.~Gong, L.~Wang, M.~Hodosh, J.~Hockenmaier, and S.~Lazebnik.
\newblock Improving image-sentence embeddings using large weakly annotated
  photo collections.
\newblock In {\em eccv}, pages 529--545. Springer, 2014.

\bibitem{HodoshYH13}
M.~Hodosh, P.~Young, and J.~Hockenmaier.
\newblock Framing image description as a ranking task: Data, models and
  evaluation metrics.
\newblock {\em Journal of Artificial Intelligence Research (JAIR)},
  47:853--899, 2013.

\bibitem{JaggiS10}
M.~Jaggi and M.~Sulovský.
\newblock A simple algorithm for nuclear norm regularized problems.
\newblock In {\em International Conference on Machine Learning (ICML)}, 2010.

\bibitem{JiICML2009}
S.~Ji and J.~Ye.
\newblock An accelerated gradient method for trace norm minimization.
\newblock In {\em International Conference on Machine Learning (ICML)}, 2009.

\bibitem{Jia13Caffe}
Y.~Jia.
\newblock {Caffe}: An open source convolutional architecture for fast feature
  embedding.
\newblock \url{http://caffe.berkeleyvision.org/}, 2013.

\bibitem{KarpathyCVPR2015}
A.~Karpathy and L.~Fei-Fei.
\newblock Deep visual-semantic alignments for generating image descriptions.
\newblock In {\em Proc. IEEE Conference on Computer Vision and Pattern
  Recognition (CVPR)}, 2015.

\bibitem{Koo08}
T.~Koo, X.~Carreras, and M.~Collins.
\newblock Simple semi-supervised dependency parsing.
\newblock {\em ACL-08: HLT}, page 595, 2008.

\bibitem{KulkarniCVPR2011}
G.~Kulkarni, V.~Premraj, S.~Dhar, S.~Li, Y.~Choi, A.~C. Berg, and T.~L. Berg.
\newblock Baby talk: Understanding and generating image descriptions.
\newblock In {\em Proc. IEEE Conference on Computer Vision and Pattern
  Recognition (CVPR)}, 2011.

\bibitem{KuznetsovaTACL2014}
P.~Kuznetsova, V.~Ordonez, T.~Berg, and Y.~Choi.
\newblock Treetalk: Composition and compression of trees for image
  descriptions.
\newblock {\em Transactions of the Association for Computational Linguistics},
  2014.

\bibitem{Mikolov2013}
T.~Mikolov, K.~Chen, G.~Corrado, and J.~Dean.
\newblock Efficient estimation of word representations in vector space.
\newblock {\em arXiv preprint arXiv:1301.3781}, 2013.

\bibitem{freeling}
L.~Padro and E.~Stanilovsky.
\newblock Freeling 3.0: Towards wider multilinguality.
\newblock In {\em Proc. Language Resources and Evaluation Conference (LREC)},
  2012.

\bibitem{Shalev-ShwartzSSC11}
S.~Shalev-Shwartz, Y.~Singer, N.~Srebro, and A.~Cotter.
\newblock Pegasos: primal estimated sub-gradient solver for svm.
\newblock {\em Mathematical Programming}, 127(1):3--30, 2011.

\bibitem{SocherTACL2014}
R.~Socher, A.~Karpathy, Q.~V. Le, C.~D. Manning, and A.~Y. Ng.
\newblock Grounded compositional semantics for finding and describing images
  with sentences.
\newblock {\em Transactions of the Association of Computational Linguistics
  (TACL)}, 2:207--218, 2014.

\bibitem{Tackstrom2012}
O.~T{\"a}ckstr{\"o}m, R.~McDonald, and J.~Uszkoreit.
\newblock Cross-lingual word clusters for direct transfer of linguistic
  structure.
\newblock In {\em Proceedings of the 2012 Conference of the North American
  Chapter of the Association for Computational Linguistics: Human Language
  Technologies}, pages 477--487. Association for Computational Linguistics,
  2012.

\bibitem{Turian2010}
J.~Turian, L.~Ratinov, and Y.~Bengio.
\newblock Word representations: a simple and general method for semi-supervised
  learning.
\newblock In {\em Proceedings of the 48th annual meeting of the association for
  computational linguistics}, pages 384--394. Association for Computational
  Linguistics, 2010.

\bibitem{VinyalsCVPR2015}
O.~Vinyals, A.~Toshev, S.~Bengio, and D.~Erhan.
\newblock Show and tell: A neural image caption generator.
\newblock In {\em Proc. IEEE Conference on Computer Vision and Pattern
  Recognition (CVPR)}, 2015.

\bibitem{weston}
J.~Weston, S.~Bengio, and N.~Usunier.
\newblock Large scale image annotation: Learning to rank with joint word-image
  embeddings.
\newblock In {\em Proc. European Conference on Computer Vision (ECCV)}, 2010.

\bibitem{XuCVPR2014}
J.~Xu, A.~G. Schwing, and R.~Urtasun.
\newblock Tell me what you see and i will show you where it is.
\newblock In {\em Proc. IEEE Conference on Computer Vision and Pattern
  Recognition (CVPR)}, 2014.

\end{thebibliography}
\end{document}